\documentclass[10pt,twocolumn,letterpaper]{article}

\usepackage{cvpr}
\usepackage{times}
\usepackage{epsfig}
\usepackage{graphicx}
\usepackage{amsmath}
\usepackage{amssymb}
\usepackage{xcolor} 
\usepackage{url}
\usepackage{amsthm}
\usepackage{amssymb}  
\usepackage{subfiles}
\usepackage{balance}

\newcommand{\PreserveBackslash}[1]{\let\temp=\\#1\let\\=\temp}
\newcommand{\norm}[1]{\lVert#1\rVert}
\newcommand{\rom}[1]{\uppercase\expandafter{\romannumeral#1}}
\def\highest #1{\textbf{\textcolor[rgb]{0.725,0.274,0.274}{#1}}}
\def\sechighest #1{\textbf{\textcolor[rgb]{0,0.494,0.278}{#1}}}
\usepackage{graphicx}
\usepackage{subfig}
\usepackage[linesnumbered, algoruled]{algorithm2e}
\usepackage{multirow}
\usepackage{array}
\usepackage{threeparttable}
\usepackage{tabularx, booktabs}

\def \fk#1 {\mathcal{F}(\boldsymbol{\theta}^{#1},\boldsymbol{G}^{#1}, \boldsymbol{P}^{#1} )}
\def \lk#1 {\mathcal{L}(\boldsymbol{\theta}^{#1},\boldsymbol{G}^{#1}, \boldsymbol{P}^{#1})}

\usepackage[pagebackref=true,breaklinks=true,letterpaper=true,colorlinks,bookmarks=false]{hyperref}

 \cvprfinalcopy 

\def\cvprPaperID{4381} 

\ifcvprfinal\fi
\begin{document}

\title{Deep Robust Subjective Visual Property Prediction in Crowdsourcing}

\author{\parbox{16cm}{\centering
	{\large  \ \ \ \ \ \ \ \ \ \ \ \ Qianqian Xu$^{1}$ \ \ \ \ \ \ \ \ \ \ \ \ \ \  Zhiyong Yang$^{2,3}$ \ \ \ \ \ \  Yangbangyan Jiang$^{2,3}$\\ 
	Xiaochun Cao$^{2,3}$ \ \ \ \ \ \ \ \  Qingming Huang$^{1,4,5}$ \ \ \ \ \ \ \ \  Yuan Yao$^{6}$ }\\
    {\normalsize
    $^1$ Key Lab of Intell. Info. Process., Inst. of Comput. Tech., CAS, Beijing, 100190, China\\
    $^2$ State Key Laboratory of Info. Security (SKLOIS), Inst. of Info. Engin., CAS, Beijing, 100093, China\\
    $^3$ School of Cyber Security, University of Chinese Academy of Sciences, Beijing,100049, China \\
    $^4$ School of Computer Science and Tech., University of Chinese Academy of Sciences, Beijing, 101408, China \\
    $^5$ BDKM, University of Chinese Academy of Sciences, Beijing, 100190, China\\
    $^6$ Department of Mathematics, Hong Kong University of Science and Technology, Hong Kong\\
    }
    {\tt\small xuqianqian@ict.ac.cn\quad\quad\{yangzhiyong,jiangyangbangyan,caoxiaochun\}@iie.ac.cn\quad\quad qmhuang@ucas.ac.cn \quad\quad yuany@ust.hk 
    }
    }
    }

\maketitle

\begin{abstract}
    
\noindent The problem of estimating subjective visual properties
(SVP) of images (e.g., Shoes A is
more comfortable than B) is gaining rising attention. Due to its highly subjective nature, different annotators often exhibit different interpretations
of scales when adopting absolute value tests. Therefore, recent investigations turn to collect pairwise comparisons via crowdsourcing platforms. However,  
crowdsourcing data usually contains outliers. For this purpose, it is desired to develop a robust model for learning SVP from crowdsourced noisy annotations. In this paper, we construct a deep SVP prediction model which not only leads to better detection of annotation outliers but also enables
learning with extremely sparse annotations. Specifically, we construct a comparison multi-graph based on the collected annotations, where different labeling results correspond to edges with different directions between two vertexes. Then, we propose a generalized deep probabilistic
framework which consists of an SVP prediction module
and an outlier modeling module that work collaboratively
and are optimized jointly. Extensive experiments on various benchmark datasets demonstrate that our new approach guarantees promising results.

\end{abstract}


%

	\maketitle

\section{Introduction}

In recent years, estimating subjective visual properties (SVP) of images \cite{fu2015robust,kovashka2017attributes,squalli2018deep} is gaining rising attention in computer vision community. SVP measures a user's subjective perception and
feeling, with respect to a certain property in images/videos. For example, estimating properties of
consumer goods such as shininess of shoes \cite{fu2015robust} improves
customer experiences on online shopping websites; and estimating
interestingness \cite{fu2014interestingness} from images/videos would be helpful for media-sharing websites (e.g., Youtube). Measuring and ensuring good estimation of SVP is thus highly subjective in nature. Traditional methods usually adopt absolute value to specify a rating from
1 to 5 (or, 1 to 10) to grade the property of a stimulus. For example, in image/video interestingness
prediction, 5 being
the most interesting, 1 being the least interesting. However, since by definition these properties are
subjective, different raters often exhibit different interpretations of the scales and as a result the annotations
of different people on the same sample can vary
hugely. Moreover, it is unable to concretely define the concept of
scale (for example, what a scale 3 means for an image), especially without any common reference point.
Therefore,
recent investigations
turn to an alternative approach with pairwise comparison. In a pairwise comparison test, an individual is simply
asked to compare two stimuli simultaneously, and votes
which one has the stronger property based on his/her perception. Therefore individual decision process in pairwise comparison
is simpler than in the typical absolute value tests, as the multiple-scale
rating is reduced to a dichotomous choice. It not only promises assessments that are easier
and faster to obtain with less demanding task for raters, but
also yields more reliable feedback with less personal scale bias
in practice. However, a shortcoming of pairwise comparison
is that it has more expensive sampling complexity than the
absolute value tests, since the number of pairs grows quadratically with
the number of items to be ranked. 

With the growth of crowdsourcing \cite{Branson_2017_CVPR}
platforms such as MTurk, InnoCentive, CrowdFlower,
CrowdRank, and AllOurIdeas, recent studies
thus resort to using crowdsourcing tools to tackle the cost problem.
However, since the participants in the crowdsourcing experiments often work in the absence of supervision, it is hard
to guarantee the annotation quality in general \cite{daniel2018quality}. If the experiment lasts too long, the raters always lose their patience and
end the test in a hurry with random annotations. Worse,
the bad users might even provide wrong answers deliberately to corrupt the system. Such
contaminated decisions are useless and may deviate significantly
from other raters' decisions thus should be identified
and removed in order to achieve a robust SVP prediction result.

Therefore, existing approaches on SVP prediction are often split into
two separate steps: the first is a standard
outlier detection problem (e.g., majority voting) and the second is a regression
or learning to rank problem. However, it
has been found that when pairwise local rankings are
integrated into a global ranking, it is possible to detect
outliers that can cause global inconsistency and yet are
locally consistent, i.e., supported by majority votes \cite{Hodge}. To overcome this limitation, \cite
{fu2015robust} proposes a more principled way to identify annotation outliers by formulating
the SVP prediction task as a unified robust learning
to rank problem, tackling both the outlier detection and SVP
prediction tasks jointly. Different from this work which only enjoys the limited representation power of the image low-level features, our goal in this paper is to leverage the
strong representation power of deep neural networks
to explore the SVP prediction issue from a deep perspective.

When it comes to deep learning, it is known that several kinds of factors can drive the deep learning model away from a perfect one, with the data perturbation issue as an typical example. Besides the notorious issue coming from the crowdsourcing process, deep learning is in itself known to be more vulnerable to contaminated data since the extremely high model complexity brings extra risks to overfit the noisy/contaminated data \cite{lu2017learning,han2018progressive,yao2018deep,zhang2018deep,jindal2016learning,DBLP:conf/nips/Vahdat17,Patrini_2017_CVPR}. We believe that how to guarantee the robustness is one of the biggest challenges when constructing deep SVP prediction models. In this sense, we propose a deep robust model for learning SVP from crowdsourcing.  As an overall summary, we list our main contributions as follows:
\begin{itemize}
    \item A novel method for robust prediction of SVP is proposed. To the best of our knowledge, our framework offers the first attempt to carry out the prediction
procedure with automatic detection of sparse outliers from a deep perspective.

\item In the core of the framework lies the unified probabilistic model, which is used to formulate the generating process of the labels when outliers exist. Based on this model, we then propose a Maximum A Posterior (MAP) based objective function.

	\item An alternative optimization scheme is adopted to solve the corresponding model. Specifically, the network parameters could be updated from the gradient-based method with the back-propagation, whereas the outlier pattern could be solved from an ordinal gradient descent method or a proximal gradient method.  
 
\end{itemize}

\section{Related Work}
\subsection{Subjective visual properties} 
Subjective visual property
prediction has gained rising attention in the last several years. It covers a large variety of computer vision
problems, including image/video interestingness \cite{fu2014interestingness}, memorability \cite{jing2017predicting}, and quality of experience \cite{MM13} prediction, etc. When used as
a semantically meaningful representation, the subjective
visual properties are often referred to as relative
attributes \cite{YangZXYHG16,kovashka2017attributes}. The original SVP prediction approach treats this task as a learning-to-rank problem. The main idea is to use ordered pairs of training images to train a ranking function that will generalize to new images. Specifically, a set of pairs ordered according to their perceived property strength is obtained from human annotators, and a ranking function that preserves those orderings is learned. Given a new image pair, the ranker indicates which image has the property more. 
A naive way to learn the ranker is to resort to traditional pairwise learning-to-rank methods such as RankSVM \cite{joachims2009svm}, RankBoost \cite{freund2003efficient}, and RankNet \cite{burges2005learning}, etc. However, these methods are not a natural fit in the scenarios with crowdsourced outliers. In \cite{fu2015robust}, it proposes a unified robust
learning to rank (URLR) framework to solve jointly
both the outlier detection and learning to rank problems. Different from this line of research, we study the robust SVP prediction in the context of deep learning. Equipped with better feature representation power,  
we show both
theoretically and experimentally that by solving both
the outlier detection and ranking prediction problems
jointly in a deep framework, we achieve better outlier detection and better ranking prediction.

\begin{figure*}[t] 
	\centering
	\subfloat{
		\includegraphics[width =\textwidth]{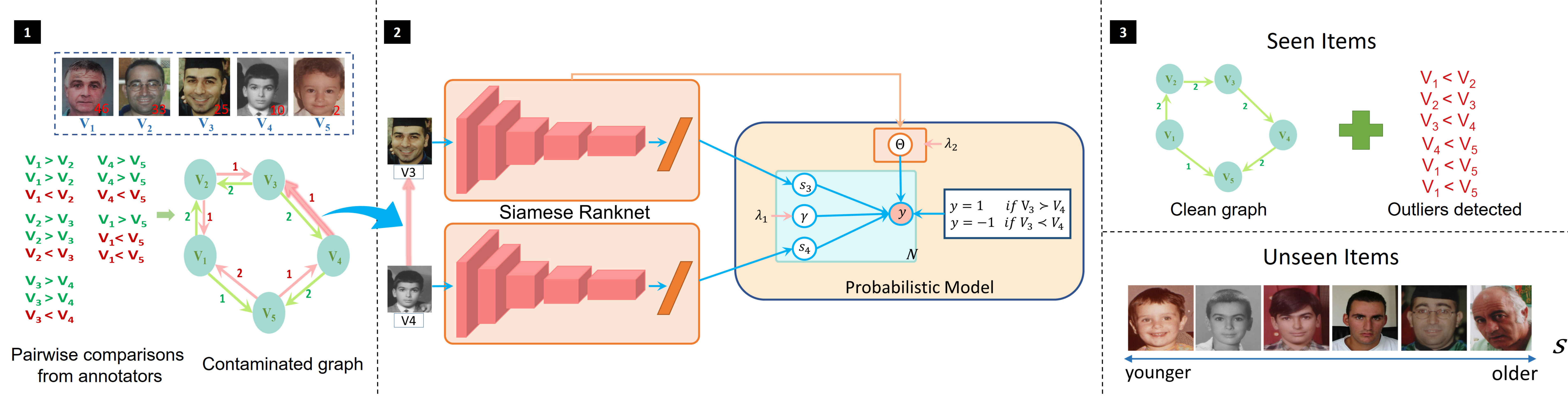}
	}
	\caption{Overview of our approach. (1) Constructing a comparison graph from the crowdsourcing annotations, which is contaminated with outlier labels. (2) We propose a generalized deep probabilistic
framework, where an outlier indicator $\boldsymbol{\gamma}$ is learned along with the network parameters $\Theta$. (3) Our Framework will output a clean graph on the training set, where contaminated annotations are eliminated. Furthermore, our model could predict a rank-preserved score for each unseen instance. Best viewed in color.}
	\label{fig:overview}
\end{figure*}

\subsection{Learning with noisy data}
Learning from noisy data has been studied extensively in recent years. Traditionally, such methods could be tracked back to statistical studies such as Majority voting, $M$-estimator ~\cite{Huber81}, Huber-LASSO \cite{MM13}, and Least Trimmed Squares (LTS)~\cite{xu2017exploring}, etc. However, these work do not have
prediction (especially with the power of deep
learning) ability for unseen samples.
Recently, there is a wave to explore robust methods to learn from noisy labels, in the context of deep learning. Generally speaking, there are four types of existing methods: (\textbf{\rom{1}}) robust learning based on probabilistic graphical models where the noisy patterns are often modeled as latent variables \cite{yao2018deep,DBLP:conf/nips/Vahdat17}; (\textbf{\rom{2}}) progressive and self-paced learning, where easy and clean examples are learned first, whereas the hard and noisy labels are progressively considered \cite{han2018progressive}; (\textbf{\rom{3}}) loss-correction methods, where the loss function is corrected iteratively \cite{Patrini_2017_CVPR}; (\textbf{\rom{4}}) network architecture-based method, where the noisy patterns are modeled with specifically designed modules \cite{jindal2016learning}. Meanwhile, there are also some efforts on designing deep robust models for specific tasks and applications: \cite{lu2017learning} proposes a method to learn from Weak and Noisy Labels
for Semantic Segmentation; \cite{zhang2018deep} proposes a deep robust unsupervised method for saliency detection, etc. 

Compared with these recent achievements, our work differs significantly in the sense that: a) We provide the first trial to explore the deep robust learning problem in the context of crowdsourced SVP learning. b) We adopt a pairwise learning framework, whereas the existing work all adopt instance-wise frameworks.

\section{Methodology}
\subsection{Problem definition}
Our goal in this paper is two-fold:
\begin{itemize}
	\item[\textbf{(a)}] We aim to learn a deep SVP
	prediction model from a set of sparse and noisy pairwise
	comparison labels. Specifically the ranking patterns should be preserved. 
	\item[\textbf{(b)}] To guarantee the quality of the model, we expect that all the noisy annotations could be detected and removed along with the training process. 
\end{itemize}

 We denote the id of two images in the $i$th pair as $i_1$ and $i_2$, and denote the corresponding image pair as ($\boldsymbol{x}_{i_1}$, $\boldsymbol{x}_{i_2}$). More precisely, we are given a pool with $n$ training images and a set of SVPs. In addition, for each SVP, we are given a set of pairwise comparison labels. Such pairwise comparison data can be represented by a directed multi-graph where multiple edges could be found between two vertexes. Mathematically, we denote the graph as $\mathcal{G} = (\mathcal{V},\mathcal{E})$. $\mathcal{V}$ is the set of vertexes  which contains all the distinct image items occurred in the comparisons. $\mathcal{E}$ is the set of comparison edges. For a specific user with id $j$ and a specific comparison pair $i$ defined on two item vertexes $i_1$ and $i_2$, if the user believes that $i_1$ holds a stronger/weaker presence of the SVP, we then have an edge $(i_1,i_2,j)$/$(i_2,i_1,j)$, respectively. Equivalently we also denote this relation as $i_1 \overset{j}{\succ} i_2$/$i_2 \overset{j}{\succ} i_1$. Since multiple users take part in the annotation process, it is natural to observe multi-edges between two vertexes. Now we could denote the labeling results as a function $\mathcal{Y}: \mathcal{E}\rightarrow \{-1,1\}$. For a given pair $i$ and a rater $j$ who annotates this pair, the corresponding label is denoted as $y_{ij}$, which is defined as :
\begin{equation}
\left\{
\begin{matrix}
y_{ij} =&{\color{white}-}1, & (i_1,i_2,j) \in \mathcal{E};\\ 
y_{ij} =&-1, & (i_2,i_1,j) \in \mathcal{E}.
\end{matrix}\right.
\end{equation}
Now we present an example of the defined comparison graph. See step 1 in Figure \ref{fig:overview}. In this figure, the SVP in question is the age of the humans in the images. Suppose we have 5 images with ground truth ages (marked with red in the lower right corner of each image), we then have $\mathcal{V} =\{1,2,\cdots,5\}$. Furthermore, we have three users with id 1, 2, 3 who take part in the annotation. According to the labeling results shown in the lower left side, we have 
$\mathcal{E} =\{(1,2,1), (1,2,2),(2,1,3), \cdots,(1,5,1),(5,1,2),(5,1,3)\}$. 
As shown in this example, we would be most likely to observe both $i_1 \succ i_2$ and $i_2 \succ i_1$ for a specific pair $i$. This is mainly caused by the bad and ugly users who provide erroneous labels. For example for vertexes 1 and 2, the edge $(2,1,3)$ is obviously an abnormal annotation. With the above definitions and explanations, we are ready to introduce the input and output of our proposed model.

\noindent \textbf{Input.} The input of our deep model is the defined multi-graph $\mathcal{G}$ along with the image items, where each time a specific edge is fed to the network. \\ \textbf{Output.} As will be seen in the next subsection, our model will output the relative score $s_{i_1}$ and $s_{i_2}$ of the image pair along with an outlier indicator which could automatically remove the abnormal directions on $\mathcal{G}$. Note that learning $s_{i_1}$ and $s_{i_2}$ directly achieves our goal \textbf{(a)}, while detecting and removing outlier directions on the graph directly achieves goal \textbf{(b)}.

\subsection{A deep robust SVP prediction model}
In contrast to traditional methods, we propose a deep robust SVP ranking model in this paper. 
According to step 2 in Figure 1,  we employ a deep Siamese \cite{sim1,sim2} convolutional neural network as the ranking model to calculate the relative scores for image pairs. In this model, the input is an edge in the graph $\mathcal{G}$ together with the image pair $(\boldsymbol{x}_{i_1}, \boldsymbol{x}_{i_2})$. Each branch of the network is fed with an image and outputs the corresponding scores $s(\boldsymbol{x}_{i_1})$ and $s(\boldsymbol{x}_{i_2})$. Then we propose a robust probabilistic model based on the difference of the scores. As a note for the network architecture, we choose an existing popular CNN architecture, ResNet-50 \cite{he2016deep}, as the backbone of the Siamese network. Such residual network is equipped with shortcut connections, bringing in promising performance in image tasks.

 With the network given, we are ready to elaborate a novel probabilistic model to simultaneously prune the outliers and learn the network parameters for SVP prediction. In our model, the noisy annotations are treated as a mixture of reliable patterns and outlier patterns. More precisely, to guarantee the performance of the whole model, we expect $s(\boldsymbol{x}_{i_1}), s(\boldsymbol{x}_{i_2})$, i.e., the scores returned by the network to capture the reliable patterns in the labels. Meanwhile, we introduce an outlier indicator term $\gamma(y_{ij})$ to model the noisy nature of the annotations. During the training process, our prediction is an additive mixture of the reliable score and the outlier indicator.\\
\indent To see how the inclusion of $\boldsymbol{\gamma}$ could help us detect and remove outlier,  one should realize that, since $y_{ij}$ must be either 1 or -1, there are only two distinct values for $\gamma(y_{ij})$, with one for each direction. If we can learn a reasonable $\gamma(y_{ij})$ such that $\gamma(y_{ij})\neq 0$ only if the corresponding direction is not reliable, we can then remove the contaminated directions in $\mathcal{G}$ and obtain a clean graph. To illustrate it in an easier way, let us back to step 1 in Figure 1. According to the lower left contents, we have three annotations for pair ($V_1$, $V_2$). We have two distinct $\gamma(y_{ij})$ for these annotations: For the correct direction, we have a $\gamma(1)$ for $(1,2,1)$ and $(1,2,2)$; For the contaminated direction, we have a different gamma with value $\gamma(-1)$ for $(2,1,3)$. Now if we can learn $\gamma(y_{ij})$ in a way that $\gamma(1)=0$ and $\gamma(-1) \neq 0$, then we can easily detect the contaminated direction $(2,1)$.\\
\indent Given the clarification above, our next step is to propose a probabilistic model of the labels based on the outlier indicator $\boldsymbol{\gamma}$, the network parameters $\Theta$, and the predicted  scores $s(\cdot)$. Specifically, we model the conditional distribution of the annotations along with the prior distribution of $\boldsymbol{\gamma}$ and $\Theta$  in the following form:
\[y_{ij}\ |\ \boldsymbol{x}_{i_1},\boldsymbol{x}_{i_2},\Theta,\gamma(y_{ij}) \overset{i.i.d}{\sim}f(y_{ij},s(\boldsymbol{x}_{i,1}, \boldsymbol{x}_{i,2},\Theta) + \gamma(y_{ij})),\]
\[\gamma(y_{ij})\ |\ \lambda_1 \overset{i.i.d}{\sim} h(\gamma(y_{ij}),\lambda_1), \ \  \Theta\ |\ \lambda_2 \sim g(\Theta,\lambda_2). \]
\begin{itemize}
	\item $s(\boldsymbol{x}_{i_1},\boldsymbol{x}_{i_2},\Theta)= s(\boldsymbol{x}_{i_1},\Theta) -s(\boldsymbol{x}_{i_2},\Theta) $ is the relative score of the annotation, which will be directly learned from the deep learning model with the parameter set $\Theta$. As mentioned above, $s(\boldsymbol{x}_{i_1},\boldsymbol{x}_{i_2},\Theta)$ are expected to model the reliable pattern in the annotations.   The prior distribution of $\Theta$ is assumed to be associated with a p.d.f. (probability density function) $p(\Theta\ |\ \lambda_2)= g(\Theta,\lambda_2)$ ($\lambda_2$ is a predefined hyperparameter), which is denoted as $g$ in short.
	\item  $\gamma(y_{ij})$ is the outlier indicator which induces unreliability. Since only outliers have a nonzero indicator, we model the randomness of $\gamma(y_{ij})$ with an i.i.d sparsity-inducing prior distribution (e.g., Laplacian distribution) with the p.d.f. being $p(\gamma(y_{ij})|\lambda_1)= h(\gamma(y_{ij}),\lambda_1)$ ($\lambda_1$ denotes the hyperparameter), which is denoted as $h_{ij}$ in short.
	\item As we have mentioned above, the noisy prediction $s(\boldsymbol{x}_{i_1},\boldsymbol{x}_{i_2},\Theta) + \gamma(y_{ij})$ is an additive mixture of the reliable score and outlier indicator. 
	\item $f(y_{ij},s(\boldsymbol{x}_{i,1}, \boldsymbol{x}_{i,2},\Theta) + \gamma(y_{ij}))$ is the conditional p.d.f. of the labels, which is denoted as $f_{ij}$ in short.
\end{itemize}  

Let $\boldsymbol{\gamma} = \{\gamma(y_{ij})\}_{(i_1,i_2,j) \in\mathcal{E}}$, $\boldsymbol{y}= \{y_{ij}\}_{(i_1,i_2,j) \in\mathcal{E}} $. Now our next step is to construct a loss function for this probabilistic model.
According to the Maximum A Posterior (MAP) rule in statistics, a reasonable solution of the parameters should have a large posterior probability $P(\Theta,\boldsymbol{\gamma}\ |\ \boldsymbol{y},\boldsymbol{X},\lambda_1,\lambda_2)$. In other words, with high probability, the parameters (\textit{$\boldsymbol{\gamma},\Theta$ in our model}) should be observed after seeing the data (\textit{$\boldsymbol{y},\boldsymbol{X}$ in our model}) and the predefined hyperparameters ($\lambda_1,\lambda_2$).  This motivates us to \textit{maximize} the posterior probability in our objective function. Furthermore, to simplify the calculation of the derivatives, we adopt an equivalent form where the negative log posterior probability is \textit{minimized}:
\[\min\limits_{\Theta,\boldsymbol{\gamma}} -\log\left(P(\Theta,\boldsymbol{\gamma}\ |\ \boldsymbol{y},\boldsymbol{X},\lambda_1,\lambda_2)\right).\]
Following the Bayesian rule,  one has: 
\begin{equation*}
\begin{aligned}
	& & &\ \ \ \ \ \ \ \ \ \ \ \ \ P(\Theta,\boldsymbol{\gamma}\ |\ \boldsymbol{y},\boldsymbol{X},\lambda_1,\lambda_2)\\
	& &=&\ \ 	
	\frac{P(\boldsymbol{y}|\ \boldsymbol{X}, \Theta, \boldsymbol{\gamma})  \cdot P(\Theta|\ \lambda_1) \cdot P(\boldsymbol{\gamma}| \lambda_2) \cdot P({\boldsymbol{X}})}{\int_{\Theta}\int_{\boldsymbol{\gamma}}P(\boldsymbol{X},\ \boldsymbol{y}|\ \Theta, \boldsymbol{\gamma}) \cdot P(\Theta|\ \lambda_1) \cdot P(\boldsymbol{\gamma}| \lambda_2) d\Theta d \boldsymbol{\gamma}}.
\end{aligned}
\end{equation*}
It then becomes clear  that $P(\Theta,\boldsymbol{\gamma} | \boldsymbol{y},\boldsymbol{X},\lambda_1,\lambda_2)$ is not directly tractable. Fortunately, since $\boldsymbol{X},\boldsymbol{y}$ are given and we only need to optimize $\Theta$ and $\boldsymbol{\gamma}$, the tedious term $\frac{P(\boldsymbol{X})}{\int_{\Theta}\int_{\boldsymbol{\gamma}}P(\boldsymbol{X},\ \boldsymbol{y}|\ \Theta, \boldsymbol{\gamma}) \cdot P(\Theta|\ \lambda_1) \cdot P(\boldsymbol{\gamma}| \lambda_2) d\Theta d \boldsymbol{\gamma}}$ becomes a constant, which suggests that:
\footnotesize
\begin{equation}\label{ppr}
\begin{aligned}
& & &\ \ \ \ \ \ P(\Theta,\boldsymbol{\gamma}\ |\ \boldsymbol{y},\boldsymbol{X},\lambda_1,\lambda_2) \\
& & \propto &\ \ \prod_{(i,j) \in \mathcal{D}} p(y_{ij}\ |\ \boldsymbol{x}_{i,1}, \boldsymbol{x}_{i,2},\gamma(y_{i,j}),\Theta)\cdot p(\gamma(y_{ij})\ |\ \lambda_1)\cdot p(\Theta\ |\ \lambda_2)
\\
& &=&\ \ \prod_{(i,j)\in \mathcal{D}}g \cdot h_{ij} \cdot f_{ij}.
\end{aligned}
\end{equation}
\normalsize
where $\mathcal{D}:\{(i,j): (i_1,i_2,j)\in \mathcal{E} \ or  \ (i_2,i_1,j) \in \mathcal{E}\}$.
This implies that our loss function could be simplified as:
\[
	\min\limits_{\Theta,\boldsymbol{\gamma}} \underset{{(i,j)\in \mathcal{D}}}{\sum}-\left(\log(f_{ij}) +\log(h_{ij})\right) - \log(g).
\]

With the general framework given, we provide two specified models with different assumptions on the distributions:
\begin{itemize}
	\item  \textbf{Model A}: If the prior distribution of $\gamma(y_{ij})|\lambda_1$ is a Laplacian distribution with a zero location parameter and a  scale parameter of $\frac{1}{\lambda1}$: $Lap(0,\frac{1}{\lambda_1}) =\frac{\lambda_1}{2} \mathrm{exp}(-\frac{|\gamma|}{1/\lambda_1})$ ; the prior distribution of  $\Theta$ is an element-wise Gaussian distribution  $\mathcal{N}(0, \frac{1}{2\lambda_2})$; and $y_{ij}$ conditionally subjects to a Gaussian distribution $\mathcal{N}(s(\boldsymbol{x}_{i,1}, \boldsymbol{x}_{i,2},\Theta) + \gamma(y_{ij}),1)$, then the problem becomes:
	\begin{equation*}
	\begin{split}
	\min\limits_{\Theta,\boldsymbol{\gamma}}& \ \underset{(i,j)\in \mathcal{D}}{\sum}\frac{1}{2}(y_{ij}-s(\boldsymbol{x}_{i,1}, \boldsymbol{x}_{i,2},\Theta)-\gamma(y_{ij}))^2 +\\ &  {\color{white}\underset{(i,j)\in \mathcal{D}}{\sum}}\ \ \ \ \lambda_1 \norm{\boldsymbol{\gamma}}_1 + \lambda_2\sum_{\theta \in \Theta} \theta^2, 
	\end{split}
	\end{equation*}
	where $\norm{\boldsymbol{\gamma}}_1=\sum_{(i,j)\in \mathcal{D}} |\gamma(y_{ij})|$.
	\item \textbf{Model B}: If we adopt the same assumption as above, except that we assume that $y_{ij}$ conditionally subjects to a Logistic-like distribution, then the problem could be simplified as:
	\begin{equation*}
	\begin{split}
	&\min\limits_{\boldsymbol{\gamma},\Theta} \ \sum_{(i,j)\in \mathcal{D}}\log(1+\varDelta_{ij}) + \lambda_1\norm{\boldsymbol{\gamma}}_1+ \lambda_2\sum_{\theta \in \Theta} \theta^2,
	\end{split}
	\end{equation*}
	where $\varDelta_{ij}=\mathrm{exp}(-y_{ij}(s(\boldsymbol{x}_{i,1}, \boldsymbol{x}_{i,2},\Theta)+\gamma(y_{ij})))$.
\end{itemize}
\subsection{Optimization}
With the model and network clarified, we then introduce the optimization method we adopt in this paper. Specifically, we employ an iterative scheme where $\boldsymbol{\gamma}$ and the network parameters $\Theta$ are alternatively updated until the convergence is reached. 
\subsubsection{Fix $\boldsymbol{\gamma}$, Learn $\Theta$}\label{gam}
When fixing $\boldsymbol{\gamma}$, we see that $\Theta$ could be solved from the following subproblem:
\[\min\limits_{\Theta}\underset{(i,j)\in \mathcal{D}}{-\sum}\log(f_{ij}) -\log(g)\]
Since $\Theta$ only depends  on the network, one could find an approximated solution by updating the network.
For \textbf{Model A}, this subproblem becomes:
\[\min\limits_{\Theta}\underset{{(i,j)\in \mathcal{D}}}{\sum}\frac{1}{2}(y_{ij}-s(\boldsymbol{x}_{i,1}, \boldsymbol{x}_{i,2},\Theta)-\gamma(y_{ij}))^2 + \lambda_2\sum_{\theta \in \Theta} \theta^2. \]
Similarly, for \textbf{Model B}, we come to a subproblem in the form:
\[\min\limits_{\Theta}\sum_{(i,j)\in \mathcal{D}} \log(1+\varDelta_{ij}) + \lambda_2\sum_{\theta \in \Theta} \theta^2. \]
\subsubsection{Fix $\Theta$, Learn $\boldsymbol{\gamma}$}\label{the}
Similarly, when $\Theta$ is fixed, we could solve $\boldsymbol{\gamma}$ from: 
\[\min\limits_{\boldsymbol{\gamma}}\underset{{(i,j)\in \mathcal{D}}}{\sum}-(\log(f_{ij}) + \log(h_{ij}))\]
This is a simple model of $\boldsymbol{\gamma}$ which does not involve the network.
For \textbf{Model A}, this subproblem becomes:
\[\min\limits_{\boldsymbol{\gamma}}\underset{(i,j)\in \mathcal{D}}{\sum}\frac{1}{2}(y_{ij}-s(\boldsymbol{x}_{i,1}, \boldsymbol{x}_{i,2},\Theta)-\gamma(y_{ij}))^2 + \lambda_1\norm{\boldsymbol{\gamma}}_1. \]
It enjoys a closed-form solution with the proximal operator of $\ell_1$ norm:
\begin{equation}\label{l1solv}
\gamma(y_{ij}) = \max(|c_{ij}| - \lambda_1,0)\cdot\text{sign}(c_{ij}),
\end{equation}
where:
\begin{equation*}
c_{ij} = y_{ij}-s(\boldsymbol{x}_{i,1}, \boldsymbol{x}_{i,2},\Theta).
\end{equation*}
For \textbf{Model B}, this subproblem becomes:
\[\min\limits_{\boldsymbol{\gamma}}\sum_{(i,j)\in \mathcal{D}}\log(1+\varDelta_{ij}) + \lambda_1\norm{\boldsymbol{\gamma}}_1. \]

Generally, there is no closed-form solution for this subproblem. In this paper, we adopt the proximal gradient method \cite{fista} to find a numerical solution.\\

\section{Experiments}

In this section, experiments are exhibited on three benchmark
datasets (see Table \ref{tab:dataset}) which fall into two categories: (1)
experiments on
human age estimation from face images (Section \ref{sec:age}), which can be considered as synthetic experiments. With the ground truth
available, this set of experiments enables us to perform in-depth evaluation
of the significance of our proposed method, (2) experiments on estimating SVPs as relative attributes (Section \ref{sec:relative} and \ref{sec:shoes}).

{\renewcommand\baselinestretch{1.0}\selectfont

\begin{table} [t] 
\caption{\label{simulated}  Dataset summary.}
\centering

\newsavebox{\tablebox}

\begin{lrbox}{\tablebox}
     \begin{tabular}{l||c|c|c}
  \hline     \textbf{Dataset} &\textbf{No.Pairs} &\textbf{No.Images} &\textbf{No.Classes}  \\
   \hline FG-Net Face Age Dataset &15,000 &1002 &1 \\
   \hline LFW-10 Dataset\cite{sandeep2014relative} &29,454 &2000 &10\\
 \hline Shoes Dataset \cite{kovashka2015discovering}  &87,946  &14,658 &7 \\
 \hline
 \end {tabular}
  \end{lrbox}
\scalebox{0.7}{\usebox{\tablebox}}
       \label{tab:dataset}
\end{table}
\par}

\subsection{Human age dataset} \label{sec:age}

In this experiment, we consider age as a subjective visual property of a face. The main difference between this SVP with the other SVPs evaluated so far is that we do have
the ground truth, i.e., the person’s age when the picture
was taken. This enables us to perform in-depth evaluation
of the significance of our proposed framework.\\
\textbf{Dataset} The FG-NET \footnote{http://www.fgnet.rsunit.com/} image age dataset contains 1002 images of 82 individuals
labeled with ground truth ages ranging from 0 to 69.
The training set is composed of the images of 41 randomly
selected individuals and the rest used as the test set.
For the training set, we use the ground truth age
to generate the pairwise comparisons, with the
preference direction following the ground-truth order. To
create sparse outliers, a random subset (i.e., 20\%) of the pairwise comparisons is reversed in
preference direction. In this way, we create a paired comparison
graph, possibly incomplete and imbalanced, with 1002 nodes and 15,000 pairwise comparison samples.

{\renewcommand\baselinestretch{1.0}\selectfont

\begin{table} [t]  
\caption{\label{simulated} Experimental results on Human age dataset.}
\centering

\begin{lrbox}{\tablebox}
     \begin{tabular}{|l||lcccl|}
  \hline   Algorithm  &\textbf{ACC} &\textbf{F1} &\textbf{Prec.} &\textbf{Rec.}  &\textbf{AUC}\\
   \hline Maj-LS &.5555 &.4673 &.4369 &.5022 &.5650 \\
   LS-with $\gamma$ &.5594 &.4729 &.4414 &.5093 &.5759\\
   Maj-Logistic &.5421 &.4687 &.4264 &.5205 &.5489\\
   Logistic-with $\gamma$ &.5585 &.4743 &.4410 &.5131 &.5735\\
  Maj-RankNet {\cite{burges2005learning}}     & .5611 & .4804 & .4445 & .5227 & .5792\\
   Maj-RankBoost {\cite{freund2003efficient}}  & .5425 & .5991 & .6458 & .5587 & .4507\\
   Maj-RankSVM {\cite{joachims2009svm}}        & .5838 & .3858 & .4517 & .3367 & .5665\\
  Maj-GBDT \cite{GBDT}   &.5827 &.3880 &.4504 &.3408 &.5619\\
 Maj-DART \cite{dart}    &.5940 &.3668 &.4648 &.3029 &.5690\\
 URLR {\cite{fu2015robust}} &.5765&.4633 &.5748 &.5131 &.5762\\
 \hline \hline LS-Deep-w/o $\gamma$  &.7313 &.6694 &.6407 &.7008 &.8060\\
  Logit-Deep-w/o $\gamma$ &.7439 &.6818 &.6584 &.7070 &.8168 \\ 
LS-Deep-with $\gamma$ &\highest{.7967} &\highest{.7414} &\highest{.7323}  &\highest{.7508} &\highest{.8784}\\
Logit-Deep-with $\gamma$  & \sechighest{.7917} &\sechighest{.7370}&\sechighest{.7228} & \sechighest{.7518}&\sechighest{.8739} \\

 \hline
 \end {tabular}
  \end{lrbox}
\scalebox{0.7}{\usebox{\tablebox}}
       \label{tab:age}
\end{table}
\par}

\noindent \textbf{Competitors} {We compare our method \textbf{Model A} and \textbf{Model B} with
10 competitors}. Note that \textbf{Model A} is the least square based deep model, while \textbf{Model B} is  a logistic regression based deep model. In the following experiments, we give \textbf{Model A} an alias as \textbf{LS-Deep}, and give \textbf{Model B} an alias as \textbf{Logit-Deep}  :\\
1) \textbf{Maj-LS}: This method uses majority voting for
outlier pruning and least squares problem for learning to rank.\\
2) \textbf{LS-with $\gamma$}: To test the improvement of merely adopting the robust model, we jointly employ the  linear regression model and our proposed robust mechanism as a baseline.   \\
3) \textbf{Maj-Logistic}: This method stands for another baseline in our work, where the majority voting is adopted for label processing followed with the logistic regression. \\
4) \textbf{Logistic-with $\gamma$}: Again, to test the improvement of merely adopting the robust model, we jointly employ the logistic regression model and our proposed robust mechanism as a baseline.\\
5) \textbf{Maj-RankSVM {\cite{joachims2009svm}}}: We record the performance of RankSVM to show the superiority of the representation learning.\\ 
6) \textbf{Maj-RankNet {\cite{burges2005learning}}}: To show the effectiveness of using a deeper network, we compare our method with the classical RankNet model preprocessed by the majority voting.\\ 
7) \textbf{Maj-RankBoost {\cite{freund2003efficient}}}: Besides the deep learning framework, it is also known that the ensemble-based methods could also serve a model for hierarchical learning and representation. In this sense, we compare our method with the RankBoost model, one of the most classical ensemble method. \\
8) \textbf{Maj-GBDT \cite{GBDT}}: Gradient Boosting  Decision Tree (GBDT) has gained surprising improvements in many traditional tasks. Accordingly, we compare our methods with GBDT to show its strength.\\ 
9) \textbf{Maj-DART \cite{dart}}: Recently, the well-known drop-out trick has also been applied to ensemble-based learning, be it the DART method. We also record the performance of DART to show the superiority of our method.\\  
10) \textbf{URLR {\cite{fu2015robust}}}: URLR is a unified robust learning to rank framework which aims to tackle both the outlier detection and learning to rank jointly. We compare our algorithm with this method to show the effectiveness of using a generalized probabilistic model and a deep architecture.\\
\textbf{Ablation}: To show the effectiveness of the proposed probabilistic model, we additionally add two competitors as the ablation. Note that the key element to detect outlier is the factor $\gamma$. In this way, the ablation competitors are formed with $\gamma$ eliminated:\\
1) \textbf{LS-Deep-w/o $\gamma$}: This is a partial implementation of \textbf{LS-Deep}, where the factor $\gamma$ is removed. \\
2) \textbf{Logit-Deep-w/o $\gamma$}: This is a partial implementation of \textbf{Logit-Deep}, where the factor $\gamma$ is removed. 

\noindent \textbf{Evaluation metrics} Because the ground-truth age is available, we adopt ACC, Precision, Recall, F1-score and AUC as the evaluation metrics to demonstrate the effectiveness of our proposed method. 

\noindent \textbf{Implementation Details} For the four deep learning methods, the learning rate is set as $10^{-4}$, and $\lambda_2$ is set as $10^{-3}$. For LS-Deep-with $\gamma$, $\lambda_1$ is set as 1.2. For Logit-Deep-with $\gamma$, $\lambda_1$ is set as 0.6.  

\noindent \textbf{Comparative Results} 
In all the non-deep competitive experiments, we adopt
LBP as the low-level features. Looking at the five-metrics results in Table \ref{tab:age}, we
see that our method (marked with red and green color) consistently outperforms all the
benchmark algorithms by a significant margin. This validates
the effectiveness of our method. In particular, it can
be observed that: (1) LS-with $\gamma$ (or Logistic-with $\gamma$) is superior to Maj-LS (or Maj-Logistic) because the global outlier detection is better than local
outlier detection (i.e., Majority voting). (2) The performance of deep methods is
better than all non-deep methods, interestingly even
the ablation baseline methods without $\gamma$ give better results than traditional methods with outlier detection, 
which suggests the strong representation power of
deep neural networks in SVP prediction tasks. (3) It is worth mentioning that our proposed Deep-with $\gamma$ methods successfully exhibit roughly $5\%-8\%$ improvement on all the five-metrics than Deep-without $\gamma$ methods, demonstrating the superior outlier
detection ability of our proposed framework. (4) Our proposed two models \textbf{A} (i.e., LS-Deep-with $\gamma$) and \textbf{B} (i.e., Logit-Deep-with $\gamma$) show comparable results on this dataset, while model \textbf{A} holds the lead by a slight margin.

\begin{figure}[t]
  \centering
  \subfloat{
    \includegraphics[width =0.48\textwidth]{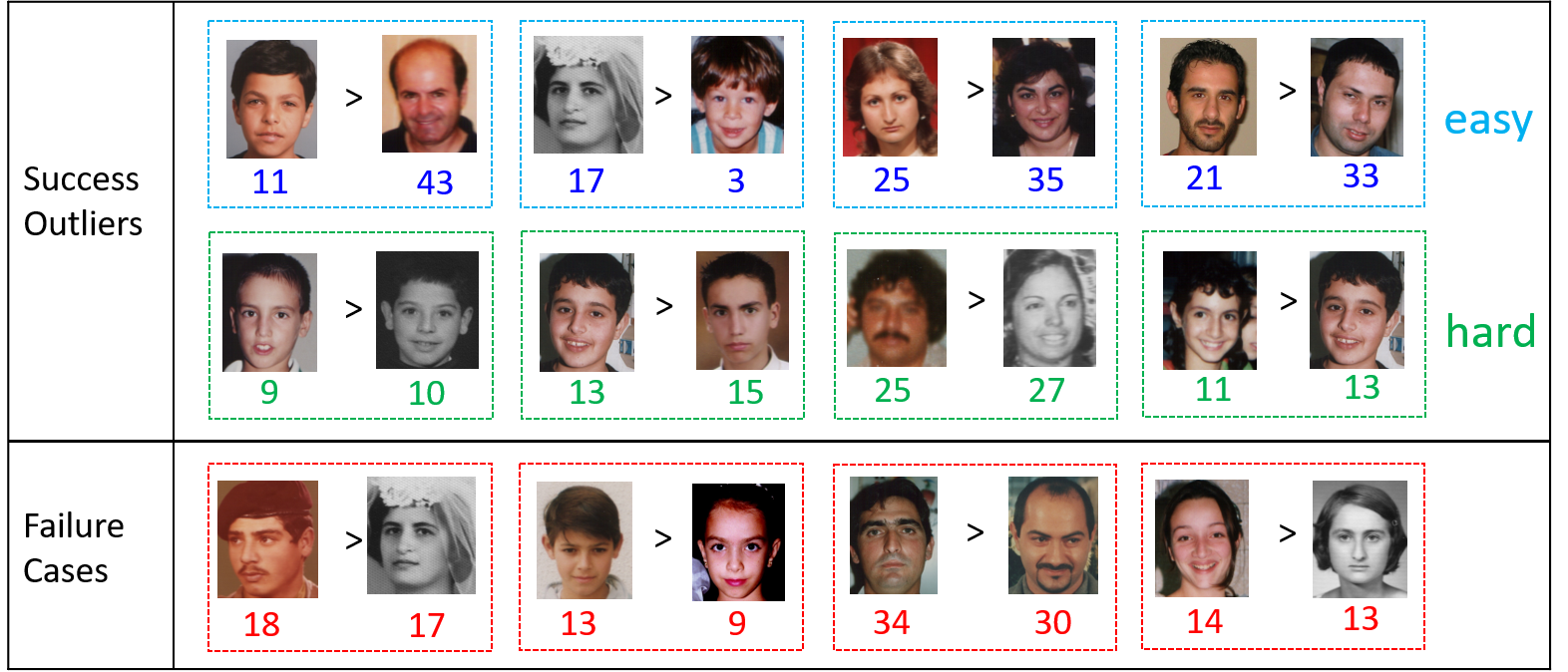}
  }
  \caption{Outlier examples detected on Human age dataset.}
  \label{fig:age_outlier}
\end{figure}

Moreover, we visualize
some examples of outliers detected by model \textbf{A}
in Figure \ref{fig:age_outlier}, while results returned by model \textbf{B} are very similar. It can be seen
that those in the blue/green boxes are clearly outliers and are
detected correctly by our method. For better illustration, the ground-truth age is printed under each image. Moreover, blue boxes show pairs with a large age differences while green boxes illustrate samples with subtle age differences, which indicates that our method not only can detect the easy pairs with a large age gap, but also can handle hard samples
with small age gap (e.g., within only 1-2 years difference). Four failure cases are shown in red boxes, in which our method treats the images on the left are older than the right one as an outlier, but the ground truth agrees with the annotation. We can easily find that this often occurs on pairs with small age differences, which indicates that our methods may occasionally lose its power when meeting highly competitive or confused pairs.

 	\renewcommand\baselinestretch{1.0}\selectfont
 	{
 	\begin{table}[t] 
		 
		\caption{\label{simulated}  Experimental results (ACC) of 10 attributes on LFW-10 dataset.}
		\footnotesize
		\centering	
 			
 		\begin{lrbox}{\tablebox}
 			\resizebox{\linewidth}{!}
 			{
	 			\begin{tabular}{|l||l@{}l@{}l@{}l@{}l@{}l@{}l@{}l@{}l@{}l@{}||c|}
	 			\hline     
	 			Algorithm &\textbf{Bald\ \ } &\textbf{D.Hai\ } &\textbf{B.Eye\ } &\textbf{GLook\ } &\textbf{Masc.\ } &\textbf{Mouth\ } &\textbf{Smile\ } &\textbf{Teeth\ } &\textbf{Foreh.\ } &\textbf{Young\ } &\textbf{Aver.}\\
	 			\hline 
	 			{Maj-LS} &.4767 &.5368 &.4787 &.4788 &.5588 &.4774 &.5220 &.5073 &.4759 &.5162 &.5029\\ 
	 			{LS-with $\gamma$}    &.5805 &.6400 &.5506 &.5932 &.6009 &.5097 &.5178 &.5198 &.5680 &.5911 &.5672\\ 
	 			{Maj-Logistic}   &.6123 &.6716 &.5146 &.5890 &.6253 &.5032 &.5031 &.5322 &.5724 &.6599 &.5784\\
	 			{Logistic-with $\gamma$}  &.6059 &.6400 &.5640 &.6038 &.6275 &.5269 &.5073 &.5405 &.5724 &.6437 &.5832 \\
	 			{Maj-RankNet {\cite{burges2005learning}}}     &.6123 &.6421 &.5551 &.6208 &.6275 &.5097 &.5304 &.5468 &.5899 &.6275 &.5862 \\ 
	 			{Maj-RankBoost {\cite{freund2003efficient}}}   &.5996 &.7053 &.5236 &.5975 &.6231 &.5097 &.5199 &.5094 &.6053 &.6032 &.5797 \\ 
	 			{Maj-RankSVM {\cite{joachims2009svm}}}        &.4852 &.6526 &.4180 &.5805 &.5588 &.4882 &.5283 &.5156 &.5482 &.6397 &.5415 \\ 
	 			{Maj-GBDT \cite{GBDT}}  &.5551 &.6253 &.4899 &.5466 &.5721 &.4903 &.5094 &.5198 &.5965 &.6235 & .5528 \\ 
	 			{Maj-DART \cite{dart} } &.5508 &.6337 &.4899 &.5339 &.5698 &.4989 &.5597 &.5364 &.5943 &.6134 & .5581 \\ 
	 			{URLR {\cite{fu2015robust}}}  &.5889    &.6538    &.6505    &.5258    &.5614    &.6319    &.5311    &.4968    &.5446    &.5570  &.5742 \\
	 			\hline \hline
	 			{LS-Deep-w/o $\gamma$ }&.5932   &.7095   &.5551   &.6081   &.5543   &.5742   &.6436   &.6133   &.5746   &.6741  &.6100  \\ 
	 			{Logit-Deep-w/o $\gamma$} &.5551 &.6758 &.5124 &.6335 &.6253 &.5806 &.6038 &.6175 &.5724 & .6235 &.6000 \\
	 			{LS-Deep-with $\gamma$ }&\highest{.6335}  &\highest{.7684}  &\highest{.5551}   &\highest{.6377}   &\highest{.6253}   &\highest{.7312}   &\highest{.7421}   &\highest{.7547}   &\highest{.6469}   &\highest{.7308} &\highest{.6826}  \\ 
	 			{Logit-Deep-with $\gamma$ }& \sechighest{.6631} & \sechighest{.7726} & \sechighest{.5798} & \sechighest{.6419} & \sechighest{.5965} & \sechighest{.7032} & \sechighest{.7358} & \sechighest{.7069} & \sechighest{.6075} & \sechighest{.6862} & \sechighest{.6694} \\
	 		 	\hline
	 			\end{tabular}
	 		}
 		\end{lrbox}
 		\scalebox{1}{\usebox{\tablebox}}
 	    \label{tab:lwf10}
 	\end{table}
 }

\begin{figure}[t]
  \centering
  \subfloat{
    \includegraphics[width =0.46\textwidth]{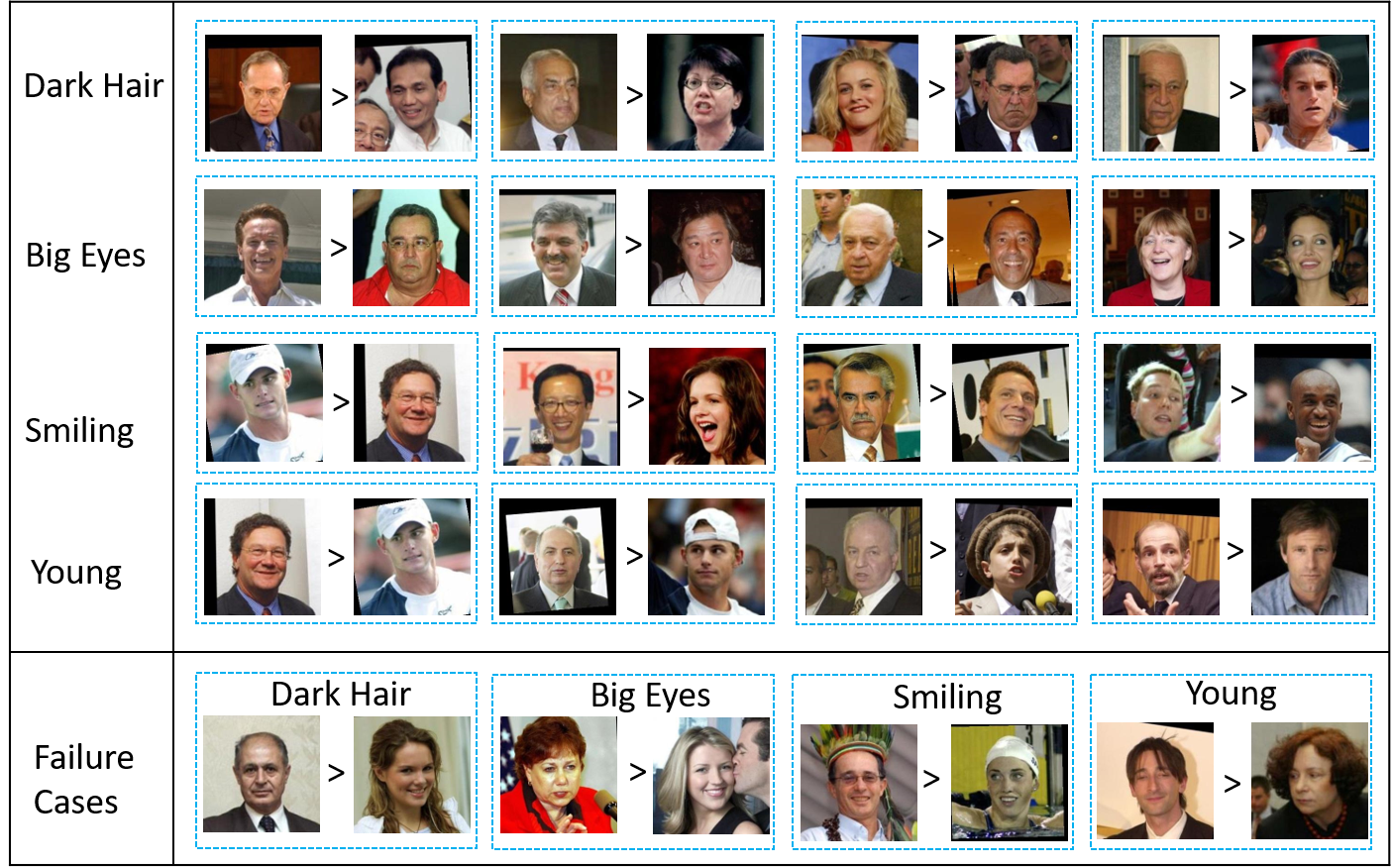}
  }
  \caption{Outlier examples of 4 representative attributes on LFW-10 dataset.}
  \label{fig:lfw_outliers}
\end{figure}

\subsection{LFW-10 dataset} \label{sec:relative}

\noindent \textbf{Dataset} The LFW-10 dataset \cite{sandeep2014relative} consists of 2,000 face images,
taken from the Labeled Faces in the Wild \cite{huang2008labeled} dataset. It
contains 10 relative attributes, like smiling, big eyes, etc. Each pair was labeled by 5
people. For example, given a specific attribute, the user will choose which one to be stronger in the attribute. As the goal of our paper is to predict SVP from noisy labels, we do not conduct any pro-precessing steps to meet the agreement of labels as \cite{just}. The resulting dataset has 29,454 total
annotated sample pairs, on average 2945 binary
pairs per attribute. 

\noindent \textbf{Implementation Details} In competitive experiments, we adopt GIST as the low-level features. 
For the four deep learning methods, the learning rate is set as $10^{-4}$, and $\lambda_2$ is set as $10^{-3}$. For LS-Deep-with $\gamma$, $\lambda_1$ is set as 1.2. For Logit-Deep-with $\gamma$, $\lambda_1$ is set as 0.5.  

\noindent \textbf{Comparative Results} Table \ref{tab:lwf10} reports the summary ACC for each attribute. The following observations can be made: (1) Our deep-methods always outperform traditional non-deep methods and ablation baseline methods for all experiment settings with higher average ACC on all attributes (0.6826 vs. 0.6100 and 0.6694 vs. 0.6000 on two models, respectively). (2) The performance of other methods is
in general consistent with what we observed in the Human age experiments.

Moreover, Figure \ref{fig:lfw_outliers} gives some examples of
the pruned pairs of 4 randomly selected attributes. In the
success cases, the left images are (incorrectly) annotated
to have more of the attribute than the right ones.
However, they are either wrong or too ambiguous to
give consistent answers, and as such are detrimental to
learning to rank. A number of failure cases (false positive
pairs identified by our models) are also shown. Some of them
are caused by unique viewpoints (e.g., for ‘dark hair’ attribute, the man has sparse scalp, so it is hard to tell who has dark hair more); others are caused
by the weak feature representation, e.g., in the ‘young’
attribute example, as ‘young’ would
be a function of multiple subtle visual cues like face shape,
skin texture, hair color, etc., whereas something like baldness
or smiling has a better visual focus captured well by 
part-based features.

 \subsection{Shoes dataset} \label{sec:shoes}

\noindent \textbf{Dataset} The Shoes dataset is collected from \cite{kovashka2015discovering} which contains 14,658 online shopping images.
In this dataset, 7 attributes are annotated by users with a wide
spectrum of interests and backgrounds. For each attribute,
there are at least 190 users who take part in the annotation,
and each user is assigned with 50 images. Note that the dataset actually uses binary annotations rather than pairwise annotations (1 for Yes, -1 for No). We then randomly sample positive annotations and negatives annotations from each user's records to form the pairs we need. For each attribute, we randomly select such 2000 distinct pairs, finally yielding a volume of 87,946 total personalized comparisons.

\noindent \textbf{Implementation Details} In competitive experiments, we concatenate the
GIST and color histograms provided by the original dataset
as the low-level features. For the LS-based deep methods, the learning rate is set as $10^{-3}$. For the Logit-based deep methods, the learning rate is set as $10^{-5}$. $\lambda_2$ is set as $10^{-3}$ for all four methods. For LS-Deep-with $\gamma$, $\lambda_1$ is set as 1.2. For Logit-Deep-with $\gamma$, $\lambda_1$ is set as 0.8.

\begin{figure}[t]
  \centering
  \subfloat{
    \includegraphics[width =0.46\textwidth]{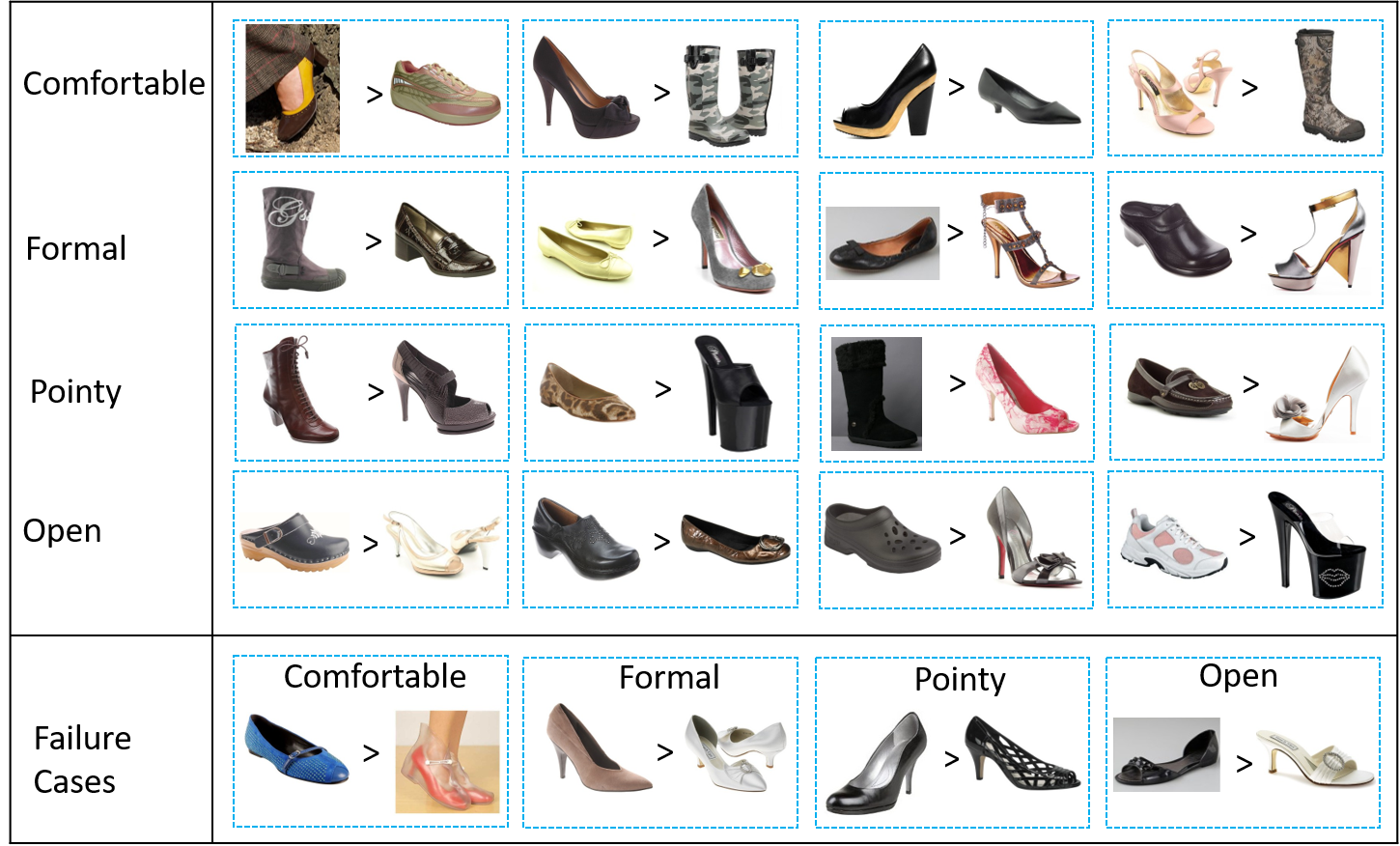}
  }
  \caption{Outlier examples of 4 representative attributes on Shoes dataset.}
  \label{fig:shoes_outliers}
\end{figure}

\noindent \textbf{Comparative Results} Similar to the Human age and LFW-10 datasets, table \ref{tab:shoes} again shows that the performance of our proposed deep models is significantly better than that of other competitors. 
Moreover, some outlier detection examples
are shown in Figure \ref{fig:shoes_outliers}. In the top four rows with successful detection
examples, the right images clearly have more of the attribute than the left ones, however are incorrectly annotated by crowdsourced raters. The failure cases are caused by the invisibility (e.g., for ‘comfortable’ attribute, though the transparent rain-boots itself is flat, there is in fact a pair of high-heeled shoes inside with red color); others are caused by different visual definitions of attributes (e.g., for ‘open’ attribute, it has multiple shades of meaning, e.g., peep-toed (open at toe) vs. slip-on (open at
heel) vs. sandal-like (open at toe and heel)); The remaining may be 
caused by ambiguity: both images have this attribute with similar degree. This thus corresponds to a truly ambiguous case which can
go either way.

{\renewcommand\baselinestretch{1.0}\selectfont

\begin{table} [t]  
\caption{\label{simulated}  Experimental results (ACC) of 7 attributes on Shoes dataset.}
\centering


\begin{lrbox}{\tablebox}
     \begin{tabular}{|l||lllllll||c|}
  \hline Algorithm  &\textbf{Comf.} &\textbf{Fash.} &\textbf{Form.} &\textbf{Pointy} &\textbf{Brown}  &\textbf{Open} &\textbf{Ornate}  &\textbf{Aver.} \\
   \hline Maj-LS 	 &.7300    &.7825    &.7325   &.7897    &.6950    &.7331    &.7300    &.7418    \\ 
   LS-with $\gamma$   &.8150    &.8125    &.7975   &.7860    &.7275    &.7444    &.7625    &.7779    \\ 
  Maj-Logistic    &.7600    &.7850    &.7475   &.7970    &.6900    &.7068    &.7175    &.7434    \\
  Logistic-with $\gamma$  &.8375    &.8175    &.7825   &.7934    &.7250    &.7444    &.7525    &.7790    \\
  Maj-RankNet {\cite{burges2005learning}}     &.7425    &.7850    &.7200   &.7860    &.6925    &.7444    &.7300    &.7429    \\ 
   Maj-RankBoost {\cite{freund2003efficient}}  &.7525    &.7300    &.7275   &.7675    &.6975    &.6955    &.6725    &.7204     \\ 
   Maj-RankSVM {\cite{joachims2009svm}}       &.7425    &.7925    &.7925   &.8081    &.6850    &.7331    &.7200    &.7534     \\ 
  Maj-GBDT \cite{GBDT}   &.7075    &.7325    &.7425   &.8007    &.6750    &.7519    &.7550    &.7379   \\ 
 Maj-DART \cite{dart}  &.6900    &.7275    &.7375   &.8376    &.6975    &.7857    &.7125    &.7412     \\ 
 URLR {\cite{fu2015robust}}  &.8200    &.8150    &.7900    &.7860    &.7325    &.7444    &.7550  &.7775\\
 \hline \hline LS-Deep-w/o $\gamma$ &.7100 &.8075 &.7400 &.7749 &.7725 &.7669 &.7050 &.7538\\ 
  Logit-Deep-w/o $\gamma$ &.7100 &.8025 &.7500 &.8044 &.7525 &.7857 &.6975 &.7575\\ 
  LS-Deep-with $\gamma$   &\highest{.8500}    &\highest{.8550}    &\highest{.8125}    &\highest{.8044}    &\highest{.8250}    &\highest{.7782}    &\highest{.8300}    &\highest{.8222}\\ 
  Logit-Deep-with $\gamma$  &\sechighest{.8550}   &\sechighest{.8500}    &\sechighest{.8200 }   &\sechighest{.8339}    &\sechighest{.8125}    &\sechighest{.7481}    &\sechighest{.8325}    &\sechighest{.8217}\\

 \hline
 \end {tabular}
  \end{lrbox}
\scalebox{0.6}{\usebox{\tablebox}}
       \label{tab:shoes}
\end{table}
\par}

\section {Conclusion}
This work explores the challenging task of SVP prediction from noisy crowdsourced annotations from a deep perspective. We present
a simple but effective general probabilistic model to simultaneously predict rank preserving scores and detect the outliers annotations, where an outlier indicator $\gamma$ is learned along with the network parameters $\Theta$. Practically, we present two specific models with different assumptions on the data distribution. Furthermore, we adopt an alternative optimization scheme to update $\gamma$ and $\Theta$ iteratively. In our empirical studies, we perform a series of experiments on three real-world datasets: Human age dataset, LFW-10, and Shoes. The corresponding results consistently show the superiority of our proposed model.   

\section{Acknowledgments}

This work was supported in part by National Basic Research Program of China (973 Program): 2015CB351800 and 2015CB85600, in part by National Natural Science Foundation of China: 61620106009, U1636214, 61861166002, 61672514, and 11421110001, in part by Key Research Program of Frontier Sciences, CAS: QYZDJ-SSW-SYS013, in part by Beijing Natural Science Foundation (4182079), in part by Youth Innovation Promotion Association CAS, and in part by Hong Kong Research Grant Council (HKRGC) grant 16303817.
 
\clearpage
\balance
{\small
\bibliographystyle{ieee}
\bibliography{egbib}
}

\end{document}